\pdfoutput=1

\documentclass[11pt]{article}

\usepackage{ACL2023}

\usepackage{times}
\usepackage{latexsym}

\usepackage{graphicx}
\usepackage{booktabs}
\usepackage{multirow}
\usepackage{comment}
\usepackage{fdsymbol}
\usepackage{accents}
\newcommand*\samethanks[1][\value{footnote}]{\footnotemark[#1]}

\usepackage[T1]{fontenc}

\usepackage[utf8]{inputenc}

\usepackage{microtype}

\title{ANALOGICAL - A Novel Benchmark for Long Text Analogy Evaluation in Large Language Models}

\author{
    Thilini Wijesiriwardene\textsuperscript{1,}\thanks{~~Corresponding author}~,   
    Ruwan Wickramarachchi\textsuperscript{1},
    Bimal G. Gajera\textsuperscript{2},\\
    \bf{Shreeyash Mukul Gowaikar\textsuperscript{3},  
    Chandan Gupta\textsuperscript{4}, 
    Aman Chadha\textsuperscript{5,6,}\thanks{~~Work does not relate to position at Amazon.}}\\
    \bf{Aishwarya Naresh Reganti\textsuperscript{7,}\samethanks~,
    Amit Sheth\textsuperscript{1},
    Amitava Das\textsuperscript{1}}\\ 
    \textsuperscript{1}AI Institute, University of South Carolina, USA,
    \textsuperscript{2}Nirma University, India, \\
    \textsuperscript{3}BITS Pilani, Goa, India,
    \textsuperscript{4}IIIT Delhi, India,
    \textsuperscript{5}Amazon AI, USA, \\
    \textsuperscript{6}Stanford, USA,
    \textsuperscript{7}Amazon, USA\\
    \texttt{thilini@sc.edu}
    }

\begin{document}
\maketitle

\begin{abstract}
Over the past decade, analogies, in the form of word-level analogies, have played a significant role as an intrinsic measure of evaluating the quality of word embedding methods such as word2vec. Modern large language models (LLMs), however, are primarily evaluated on extrinsic measures based on benchmarks such as GLUE and SuperGLUE, and there are only a few investigations on whether LLMs can draw analogies between long texts. In this paper, we present ANALOGICAL, a new benchmark to intrinsically evaluate LLMs across a taxonomy of analogies of long text with six levels of complexity -- (i) word, (ii) word vs. sentence, (iii) syntactic, (iv) negation, (v) entailment, and (vi) metaphor. Using thirteen datasets and three different distance measures, we evaluate the abilities of eight LLMs in identifying analogical pairs in the semantic vector space. Our evaluation finds that it is increasingly challenging for LLMs to identify analogies when going up the analogy taxonomy. 
\end{abstract}

\section{Introducing ANALOGIAL - a Benchmark for Analogy}
\vspace{-.1em}
The ability of humans to perceive a situation in one context as similar to that in a different context is known as \textit{analogy-making}. It is considered to be a central component of human cognition and learning.
Analogy-making has received attention from a broad audience, including cognitive scientists \cite{gentner1997structure,holyoak2001place}, linguists \cite{itkonen2005analogy}, and educators \cite{richland2015analogy} during the last several decades. Current neural network-based word embeddings are primarily influenced by the distributional hypothesis \textit{"You shall know a word by the company it keeps"} \cite{firth1957}.

\begin{figure}[!ht]
\centering
\includegraphics[width=1\columnwidth]{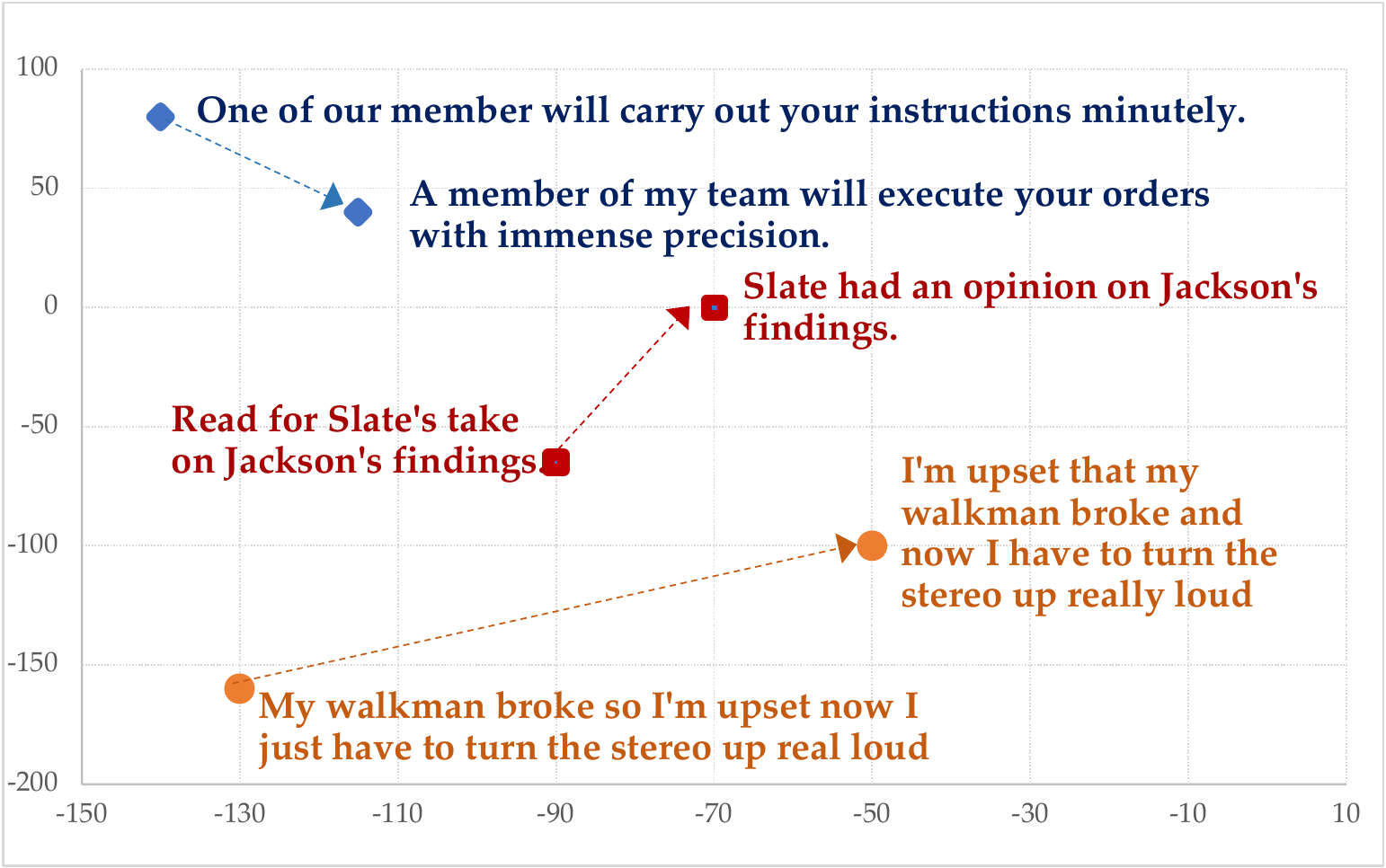}
\caption{Expected vector space embeddings of three analogical sentence pairs from a hypothetical LLM that captures sentence analogies accurately.}
\label{sent_analogy_roberta.pdf}
\end{figure}

During 2013-2017, less complex, word-level analogies played a central role in intrinsically evaluating the quality of word embedding methods, such as word2vec \cite{mikolov2013efficient}, GloVe \cite{pennington-etal-2014-glove}, and fastText \cite{bojanowski2017enriching}. Different types of textual analogies can be identified, such as word analogies \cite{gladkova-etal-2016-analogy}, proportional analogies \cite{mikolov2013efficient}, and long-text analogies \cite{ichien2020verbal}. The techniques to create word embeddings have progressed from categorical (i.e., one-hot, bag-of-words)  to continuous contextualized techniques exemplified by LLMs such as BERT \cite{devlin2018bert} and T5 \cite{10.5555/3455716.3455856}.

\begin{figure*}[!ht]
\center
\includegraphics[width=\textwidth]{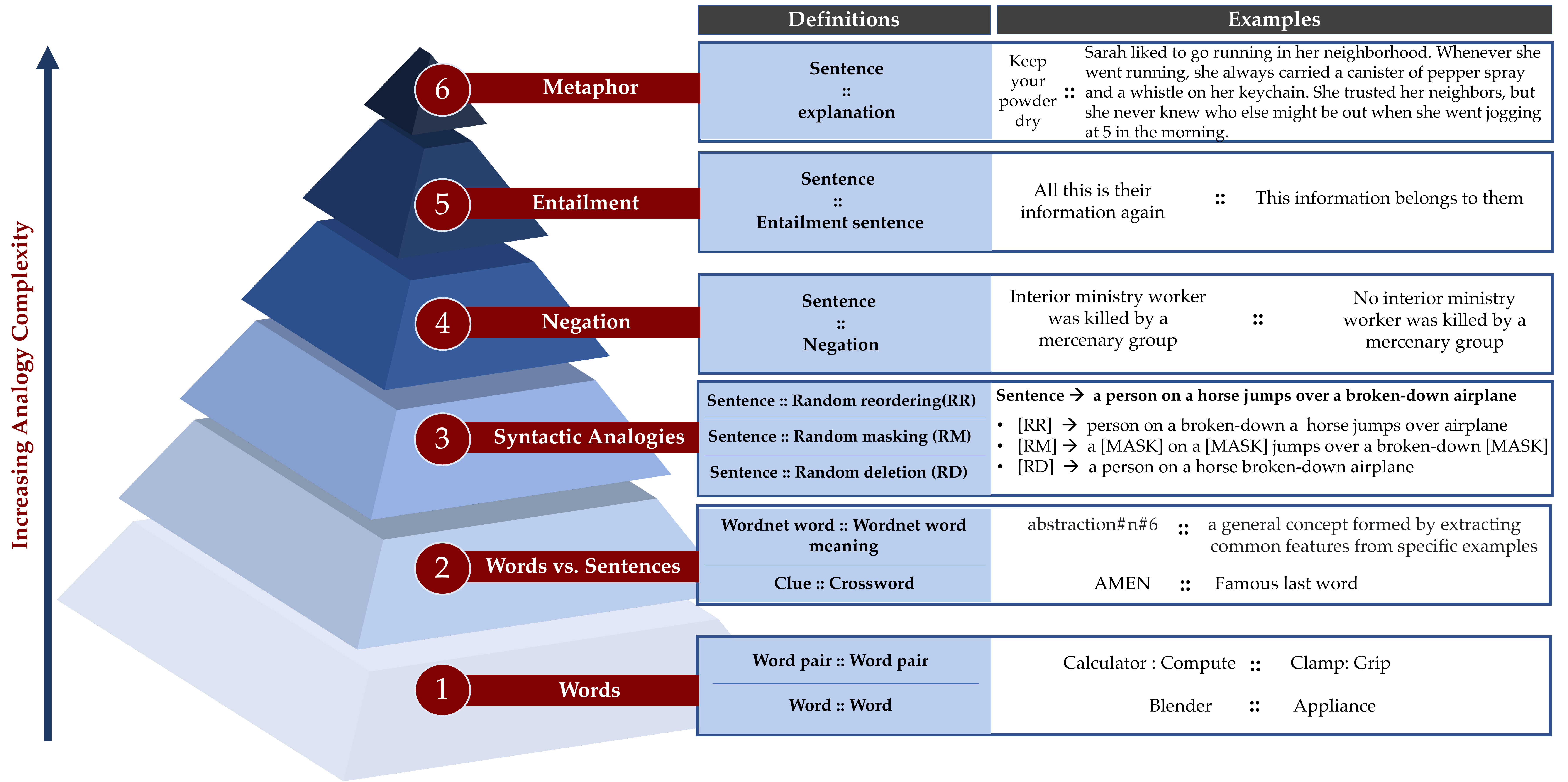}
\caption{Analogy taxonomy with six levels. The definitions of the analogies at each level and examples for each analogy type from the datasets are indicated.}
\label{analogy_tower}
\end{figure*}

However, only a few investigations have been done on the capabilities of  LLMs to draw analogies between long text \cite{czinczoll2022scientific}. For example - embeddings of sentences `\textit{I can speak two languages.}' and `\textit{I am bilingual.}' should be close-by in vector space and `\textit{I like chocolate.}' and `\textit{I do not like chocolate.}' should not be close-by. Performance evaluations of modern LLM are driven mainly by extrinsic measures based on benchmarks such as GLUE \cite{wang2018glue}, and SuperGLUE \cite{wang2019superglue}. We take this opportunity to introduce a new benchmark to \textit{intrinsically evaluate} LLMs using analogies consisting of long text (sentences, paragraphs). We hypothesize that an LLM should be able to organize the semantic vector space so that analogical lexical pairs are closer to each other (see Figure \ref {sent_analogy_roberta.pdf}).

In this paper, we introduce ANALOGICAL - a benchmark based on an analogy taxonomy consisting of six levels of analogy complexity - (i) word level, (ii) word vs. sentence level, (iii) syntactic level, (iv) negation level, (v) semantic (entailment) level and (vi) metaphor level. We proxy analogy complexity with the length of lexical items compared. We derive five and identify eight datasets at each level of the analogy taxonomy.

Euclidean distance and cosine similarity are the de facto standards for capturing analogy in the NLP community. We show that, in contrast, such measures suffer from several correlations and indirect dependencies among the vector dimensions. Finally, we argue and empirically report that Mahalanobis distance \cite{mahalanobis1936generalized} better captures the semantic equivalence in high dimensional vector spaces.

\section{Related Work}

In this section, we elaborate on previous work on analogy identification, the background of encoder-based language models and distance measures used in analogy-based comparisons in NLP.

There has been previous work on analogy identification by \citet{turney2008latent} applying singular value decomposition (SVD) \cite{golub2013matrix} based approach and by \citet{mikolov2013efficient, gladkova-etal-2016-analogy} using static word embeddings with vector offset approaches. In more contemporary literature, \citet{ushio2021bert} evaluates the ability of LMs such as BERT, GPT-2 and RoBERTa to identify word analogies in a zero-shot setting with prompts. In this work, we perform more comprehensive evaluations, including several types of analogies in addition to word analogies. We also evaluate the analogy identification abilities of eight contemporary LLMs.

Current neural network-based LMs play a pivotal role in the present-day NLP landscape by performing exceptionally well in numerous NLP tasks such as machine translation \cite{zhang2015deep, singh2017machine}, classification \cite{marwa2018deep}, and sentiment analysis \cite{hoang2019aspect}. These LMs are trained on large, heterogeneous text corpora resulting in pretrained LMs  that are then used on downstream tasks via supervised fine-tuning. This work uses the pretrained LMs in a zero-shot setting for embedding creation.

Previous research in NLP has used cosine distance/ similarity, Euclidean distance and Mahalanobis distance as popular distance measures to quantify the semantic similarity between text \cite{agarwala2021detecting, han2021survey, sunilkumar2019survey, bollegala2009measuring}. Even though Mahalanobis distance has been popularly used to measure the distance between a sample and a distribution, it has been increasingly used to measure the distance between two samples in a dataset \cite{balasubramanian2016mahalanobis, rahman2018unified}. This work extends these distance measures to measure the analogy between two lexical items.

\section{ANALOGICAL - Six Levels of Analogy}
\vspace{-.05em}

ANALOGICAL is a comprehensive benchmark focusing on six distinct categories of analogies organized within a taxonomy. These categories are determined based on the level of complexity they pose for current LLMs. Even though current language models perform exceptionally well on tasks that involve pattern recognizing the underlying text distribution and learning to infer correlations, they struggle with complex and intricate tasks such as basic symbol manipulation \cite{piekos-etal-2021-measuring}, compositionality \cite{dankers-etal-2022-paradox}, and appropriating commonsense knowledge \cite{zhou2020evaluating}. In higher levels of this taxonomy, the LMs are required to identify analogies between long and more abstract texts and, when doing so, have to face the complexities highlighted above. In the next section, we formally introduce the analogy taxonomy and the datasets representing each level in the taxonomy.

Analogies are often expressed as an explicit or implicit relational similarity, involving two main lexical items. In this work, these two lexical items vary from single words to word phrases or sentences. More formally, we denote analogy as $X :: Y$, where $X$ and $Y$ are the two lexical items and analogy is a symmetric relation.
The taxonomy of analogy is divided into six levels (see figure \ref{analogy_tower}) where complexity is increased from bottom to top.

In this section, we identify and introduce different datasets corresponding to each level of complexity in the analogy taxonomy that can be used to evaluate the performances of several SOTA language models. Table 1 summarizes the dataset statistics.
\vspace{-.2em}

\subsection{Level One}
\vspace{-.2em}

\subsubsection{Word level} 
In this level of analogy, the two analogous lexical items are either single words or word pairs. If all lexical items in a language are in set A, then the analogy between two single words $a \in W$ and $b \in W$ are denoted by $a:: b$. An analogy between two-word pairs (also known as proportional analogies) where $a, b, c, d, \in W$ is denoted by $a:b :: c:d$. This indicates that \textit{a is related to b as c is related to d}.

\subsubsection{Datasets for Level One}
This level represents word analogies. We identify four datasets at this level. Two of them, namely the \textbf{Bigger Analogy Test Set (BATS)} \cite{GladkovaDrozd2016} and \textbf{MSR Dataset} \cite{gao2014wordrep}, contain analogies between two words. 
We use the MSR dataset as is and slightly modify the BATS dataset as below for our intended use.

BATS Dataset consists of four main analogy types namely \textit{Morphology-inflections, Morphology-derivation, Semantics-encyclopedia and Semantics-lexicography}. Semantics-lexicography data contain hypernyms, hyponyms and synonyms where one word is identified to be analogous to several other words (e.g. afraid :: terrified/ horrified/ scared/ stiff/ petrified/ fearful/ panicky). In this case, we identify each element on the right as analogous to the element on the left separately (e.g., for the example above, afraid :: terrified, afraid :: horrified, etc.).

We identify two other datasets for word pair analogies in level one of the taxonomy. One is referred to as the \textbf{Google Dataset} \cite{Mikolov2013EfficientEO}, with syntactic and semantic analogies. The other comprises educational resources such as analogy problems from SAT exams (US college admission tests) and other similar problems targeted at younger students in US school system. We use these data aggregated by \citet{ushio2021bert} and identify it as the \textbf{SAT Dataset}.

\subsection{Level Two}
\subsubsection{Word vs. Sentence Level}
This level consists of analogies between a word $w$ and a sentence $S$, denoted by $S :: w$. Sentence $S$ is a sequence of words $ S = [a_1,\cdots, a_n]$ and word $w$ is $\{w_1,\cdots, w_n\} \in W$.

\subsubsection{Datasets for Level Two}
This level consists of two datasets with single words and their analogous sentences. The first dataset, \cite{pwanson_2016}, is a crossword puzzle dataset where the crosswords are words and clues are sentences/phrases (e.g., amen :: famous last words). We identify this dataset as the \textbf{Crossword Dataset}. The second dataset is the \textbf{WordNet Dataset}. WordNet is a large lexical database of English words grouped into cognitive synonym sets known as synsets \cite{miller-1992-wordnet}. The two lexical terms of interest in this dataset are the WordNet words and the different senses of these words explained in a sentence/phrase.

\subsection{Level Three}
\subsubsection{Syntactic Level}
These analogies are between single sentences. We propose that a single sentence $S$ with a word sequence $[w_1,\cdots, w_n] \in W$ is analogous to a syntactically altered version of the same sentence. We generate altered versions of original sentences by random deletion, random reordering, and random masking of the words in the sentence. If an original sentence is denoted by a word sequence $[w_1,w_2,w_3,w_4,w_5]$, an altered version of the sentence $S_{RD}$ is created by randomly deleting a consecutive range of tokens such as $[w_1,w_4,w_5]$. Another altered version is created by random reordering of the original sentence denoted by $S_{RR}$ where the altered sentence would look like $[w_1,w_2,w_4,w_3,w_5]$. The final alteration masks random words ($S_{RM}$) in the original sentence resulting in an altered version of $[w_1,[MASK],w_3,[MASK],w_5]$.

\subsubsection{Datasets for Level Three}
We are looking at analogies between two syntactically equivalent sentences at this level. We are introducing three datasets on three types of syntactic equivalence variants: random deletion, random masking, and random reordering. 
We use the sentence tagged as "neutral" in the SNLI dataset \cite{bowman2015large} as the basis for creating all three datasets introduced at this level. To create the \textbf{Random Deletion Dataset}, we delete 20\% of the words in a sentence randomly; to create the \textbf{Random Masking Dataset}, we randomly replace 20\% of tokens in a sentence with \texttt{[MASK]}. Finally, to create the \textbf{Random Reorder Dataset}, we randomly reorder 20\% of the words in a sentence. The original sentence and its altered version are identified as an analogous pair.

\subsection{Level Four}

\subsubsection{Negation Level}
The two lexical items considered in this level are single sentences, one negating the other denoted by $S$ and $S_{NG}$.

\subsubsection{Datasets for Level Four}
We identify sentences and their negated forms as a pair. Since a sentence and its negation are recognized as opposites to each other, we postulate that this is a non-analogy. We use Stanford Contradiction Corpora (specifically the negation dataset) \cite{de2008finding}. We extract the sentence with negation markers and create sentence pairs from each of these extracted sentences by keeping the negation marker and removing it. We identify this dataset as \textbf{Negation Dataset}.

\subsection{Level Five}
\subsection{Entailment Level}
This level again contains analogies between sentences. The type of analogies contained in this level is entailing sentences. Textual Entailment attempts to infer one sentence from the other. We propose that entailment considers attributional and relational similarities between sentences, making them analogous. More formally given a sentence as $S$, its entailment sentence as $S_{ET}$, words in the sentence as $w$ and words in the entailment sentence as $w'$, $S = [w_1 \cdots w_n]$, $S_{ET} = [w'_1 \cdots w'_n]$ and $S :: S_{ET}$.

\subsubsection{Datasets for Level Five}
We identify one dataset for this level and refer to it as the \textbf{Entailment Dataset}. We extract the sentence pairs tagged with the "entailment" relationship from the SNLI dataset \cite{bowman2015large} to create the data points. 

\subsection{Level Six}
\subsubsection{Metaphor Level}
This is the highest level in the taxonomy with the most complexity with regard to analogy identification, with the least attention from the NLP community. In this level, the two lexical items are a sentence and a paragraph. If a sentence is denoted by $S = [w_1 \cdots w_n]$, a paragraph is denoted by several sentences that do not include the original sentence. $P = [s_1 \cdots s_n]$. The analogy is indicated by $S :: P$.

\subsubsection{Datasets for Level Six}
We have metaphors at the top level of the analogy taxonomy. We identify two datasets at this level. One is "ePiC", a crowdsourced proverb dataset by \citet{ghosh-srivastava-2022-epic} with narratives explaining each proverb. Since the proverb and its explanation essentially have the same meaning, we assume that a proverb and its corresponding narrative are analogous to each other. We refer to this dataset as \textbf{ePiC Dataset}. Similarly, the second dataset \cite{rudrapal2017quotology} includes quotes and the elaborated meaning of each quote. We refer to this dataset as the \textbf{Quotes dataset}.

\renewcommand{\tabcolsep}{4pt}
\begin{table}
\centering
\small
\label{tab:dataset-statistics}
\begin{tabular}{ccc} 
\toprule
\begin{tabular}[c]{@{}c@{}}Levels in Analogy \\Taxonomy\end{tabular} & Dataset         & \# Datapoints  \\ 
\midrule
\multirow{4}{*}{Level One}                                           & MSR             & 44584          \\
                                                                     & BATS\textbf{*}            & 2880           \\
                                                                     & Google          & 19544          \\
                                                                     & SAT             & 1106            \\ 
\hline
\multirow{2}{*}{Level Two}                                           & Crossword       & 100000         \\
                                                                     & WordNet         & 104356         \\ 
\hline
\multirow{3}{*}{Level Three}                                         & Random Deletion\textbf{*} & 100000         \\
                                                                     & Random Masking\textbf{*}  & 100000         \\
                                                                     & Random Reorder\textbf{*}  & 100000         \\ 
\hline
Level Four                                                           & Negation\textbf{*}      & 100000         \\ 
\hline
Level Five                                                           & Entailment        & 100000         \\ 
\hline
\multirow{2}{*}{Level Six}                                           & ePiC            & 42501          \\
                                                                     & Quotes          & 998            \\
\bottomrule
\end{tabular}
\caption{Statistics of datasets used at each level of the Analogy taxonomy. Datasets derived by authors are indicated with \textbf{*}.}
\end{table}

\section{Large Language Models to Evaluate ANALOGICAL}
Modern LLMs are built upon the transformer architecture \cite{vaswani2017attention}. The LLMs we use in this study fall into two classes based on their training objective. \textbf{Masked language models (MLMs)} are trained to predict randomly masked tokens (random words replaced by a \texttt{[MASK]} token) based on all the other words present in a sequence in a bidirectional manner. MLMs use the \textit{encoder} portion of the transformer architecture. \textbf{Encoder-decoder language models(EDLMs)} build upon the entire \textit{encoder-decoder} architecture of transformers and are trained by predicting the original sequence of text given a corrupted version of the text sequence. 

\begin{figure*}[!ht]
\center
\includegraphics[width=\textwidth]{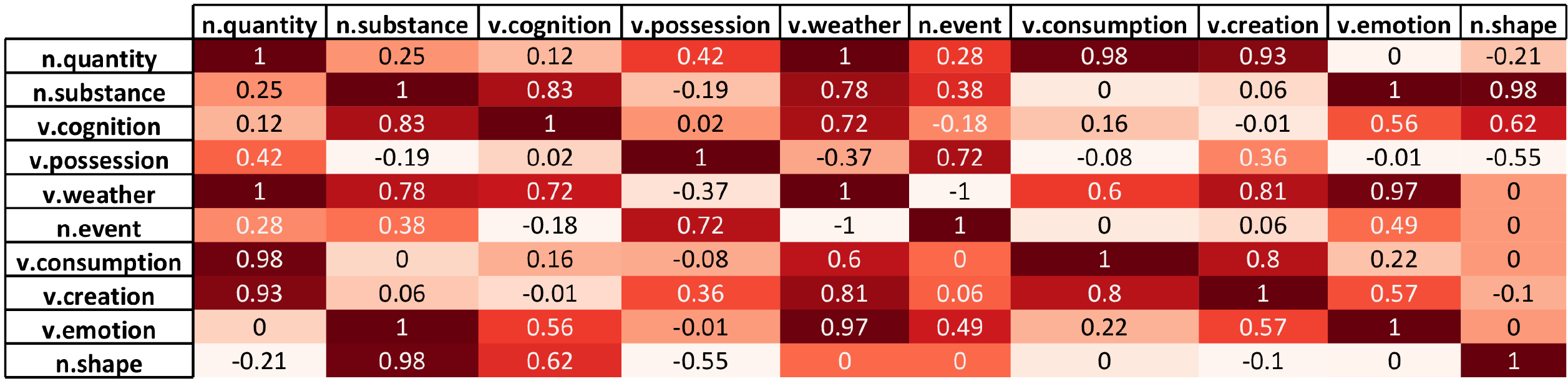}
\caption{Pearson correlation between 10 word-vector dimensions. VERB.CONSUMPTION is highly correlated with the dimension NOUN.QUANTITY and dimension of VERB.WEATHER is highly correlated with VERB.EMOTION.}
\label{fig:pearson-cor}
\end{figure*}

In the current empirical study, we examine the performance of  eight  popular pretrained language models on identifying analogies introduced in the analogy taxonomy without fine-tuning (zero-shot setting). We choose six MLM-based LLMs, namely (i) BERT \cite{devlin2018bert}, (ii) RoBERTa \cite{liu2019roberta}, (iii) AlBERT \cite{lan2019albert}, (iv) LinkBERT \cite{yasunaga2022linkbert}, (v) SpanBERT \cite{joshi2020spanbert}, and (vi) XLNet \cite{yang2019xlnet}, T5 \cite{raffel2020exploring}, an encoder-decoder-based model, and ELECTRA \cite{clark2020electra} an LLM with two transformers, one as a generator and the other as a discriminator. We include further details of these LLMs in Appendix \ref{sec:appendix-detailed-llms}).

\section{Distance Measures and Their Importance}
Previous work \cite{mikolov2013efficient, gladkova-etal-2016-analogy} used static word embeddings with vector offset approaches (such as \textit{3CosMul, 3CosAdd}) to identify word analogies.
 In this work, we use the distance between the lexical items in a high-dimensional vector space to identify the analogy between two lexical items. We identify three distance measures, namely, cosine distance (CD), Euclidean distance(ED), and Mahalanobis distance(MD). Next, we briefly explain MD. CD and ED are explained in the appendix.

 \subsection{Mahalanobis Distance (MD)}

ED does not perform well if the vector dimensions depend on each other. Mahalanobis distance \cite{mahalanobis1936generalized}, is a generalized extension of the Euclidean distance that takes into account the correlation between vector dimensions, thereby providing a balanced measure of dissimilarity. In the next section, we show that word vectors' dimensions are highly correlated. Therefore, we use MD in this work to get an accurate distance measure. Given two vectors $A = [a_i, \cdots, a_n]$ and $B = [b_i, \cdots, b_n]$, MD between the two points are given by ($C^{-1}$ indicates the covariance matrix of the dataset.):

\begin{align*}
    MD(\overrightarrow{A}, \overrightarrow{B}) &= \sqrt{(\overrightarrow{A} - \overrightarrow{B})^T C^{-1} (\overrightarrow{A} - \overrightarrow{B})}
\end{align*}
    
\subsection{Importance of Mahalanobis Distance as a Distance Measure \label{sec:importance-md}}

Vector representations of lexical items produced by LLMs are opaque due to the low interpretability of individual vector dimensions. \citet{tsvetkov2015evaluation} introduce QVEC, which uses a subspace alignment technique to align linguistic properties with distributional vector dimensions. \\
Wordnet divides verbs and nouns into 41 coarse semantic categories known as supersenses. For example, NOUN.QUANTITY and NOUN.SHAPE is supersenses related to nouns and VERB.POSSESSION and VERB.CREATION are supersenses related to verbs. SemCor is a corpus containing  13,174 noun lemmas and 5,686 verb lemmas from wordnet, and these are annotated with supersenses. Terms from SemCor are converted into linguistic word vectors based on term frequency, resulting in a set of 4,199 linguistic word vectors, each with 41 interpretable dimensions.\\
QVEC aligns distributional word vector dimensions with above described linguistically interpretable word vector dimensions through Pearson's correlations-based matrix alignments. 
We use the same methods to calculate Pearson’s correlation between the 41 vector dimensions to identify the correlations among them. Figure \ref{fig:pearson-cor} illustrates a subset of 10 vector dimensions and their correlations. We see that dimension VERB.CONSUMPTION is highly correlated with the dimension NOUN.QUANTITY and dimension of VERB.WEATHER is highly correlated with VERB.EMOTION.

Due to the correlated nature of vector dimensions, and the ability of MD to take into account the correlations between vector dimensions when calculating the distance measures, we identify MD as the best distance measure among CD, ED, and MD.

\section{Experiment Settings}
We have set up comprehensive experiments across eight LLMs, thirteen datasets, and three distance measures adding up to 312 (\texttt{$ 8 \times 13 \times 3 $}) experiments. We analyze the performance of LLMs across the analogy taxonomy by comparing the normalized distance measures. We present the complete results table for all the experiments in Appendix \ref{sec:appendix-detailed-results}).

\begin{figure*}[!ht]
\center
\includegraphics[width=\textwidth]{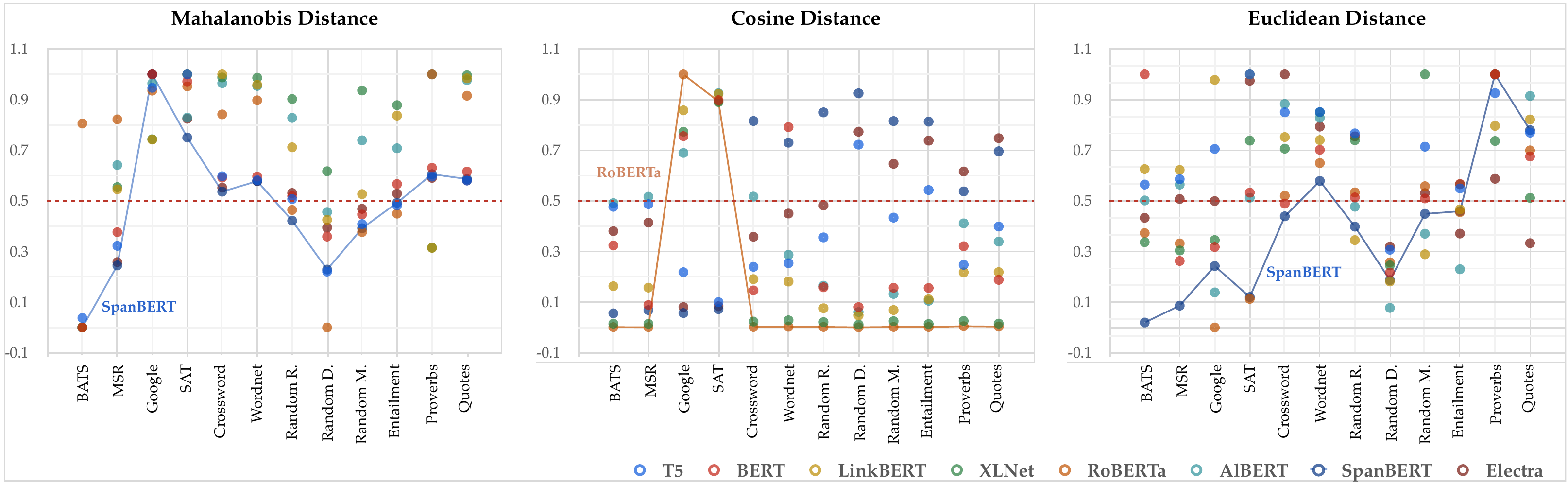}
\caption{Performance of LLMs across the thirteen datasets. All three distance measures are normalized to be in [0,1] range, 0 indicating the best performance (i.e., the least average distance between the analogous pairs). The solid lines indicate the performance of the best-performing model across all the datasets (e.g., SpanBERT outperforms the other LLMs in most datasets based on Mean MD; therefore, the line represents the fluctuations of SpanBERT's performance across the datasets).}
\label{fig:results-summary}
\end{figure*}

\begin{table*}[!ht]
\centering
\scriptsize
\begin{tabular}{lrrrrrrrrrrrrr} 
\toprule
{\begin{tabular}[c]{@{}l@{}}Language \\Model\end{tabular}}  & \multicolumn{1}{l}{BATS\_3.0} & \multicolumn{1}{l}{MSR} & \multicolumn{1}{l}{Google} & \multicolumn{1}{l}{SAT} & \multicolumn{1}{l}{Crossword} & \multicolumn{1}{l}{Wordnet} & \multicolumn{1}{l}{\begin{tabular}[c]{@{}l@{}}Random \\Reordering\end{tabular}} & \multicolumn{1}{l}{\begin{tabular}[c]{@{}l@{}}Random \\Deletion\end{tabular}} & \multicolumn{1}{l}{\begin{tabular}[c]{@{}l@{}}Random \\Masking\end{tabular}} & \multicolumn{1}{l}{\begin{tabular}[c]{@{}l@{}} \\Negation\end{tabular}} & \multicolumn{1}{l}{Entailment} & \multicolumn{1}{l}{\begin{tabular}[c]{@{}l@{}}Proverbs\\(Epic)\end{tabular}} & \multicolumn{1}{l}{Quotes}  \\ 
\midrule
T5             & 0.04                          & 0.32                    & 0.95                       & 1.00                    & 0.60                          & 0.58                        & 0.51                                                                            & 0.22                                                                          & 0.41                                                                         & 0.00                                                                            & 0.48                           & 0.60                                                                         & 0.58                        \\
BERT           & 0.00                          & 0.38                    & 1.00                       & 0.97                    & 0.59                          & 0.60                        & 0.52                                                                            & 0.36                                                                          & 0.45                                                                         & 0.37                                                                            & 0.57                           & 0.63                                                                         & 0.62                        \\
LinkBERT       & 0.00                          & 0.55                    & 0.74                       & 1.00                    & 1.00                          & 0.96                        & 0.71                                                                            & 0.43                                                                          & 0.53                                                                         & 0.46                                                                            & 0.84                           & 0.31                                                                         & 0.98                        \\
XLNet          & 0.00                          & 0.55                    & 0.74                       & 1.00                    & 0.99                          & 0.99                        & 0.90                                                                            & 0.62                                                                          & 0.94                                                                         & 0.60                                                                            & 0.88                           & 0.32                                                                         & 1.00                        \\
RoBERTa        & 0.81                          & 0.82                    & 0.94                       & 0.95                    & 0.84                          & 0.90                        & 0.46                                                                            & 0.00                                                                          & 0.38                                                                         & 0.00                                                                            & 0.45                           & 1.00                                                                         & 0.92                        \\
AlBERT         & 0.00                          & 0.64                    & 0.96                       & 0.83                    & 0.97                          & 0.95                        & 0.83                                                                            & 0.46                                                                          & 0.74                                                                         & 0.51                                                                            & 0.71                           & 1.00                                                                         & 0.98                        \\
SpanBERT       & 0.00                          & 0.25                    & 1.00                       & 0.75                    & 0.54                          & 0.58                        & 0.42                                                                            & 0.23                                                                          & 0.39                                                                         & 0.23                                                                            & 0.49                           & 0.61                                                                         & 0.59                        \\
Electra        & 0.00                          & 0.26                    & 1.00                       & 0.82                    & 0.55                          & 0.58                        & 0.53                                                                            & 0.39                                                                          & 0.47                                                                         & 0.34                                                                            & 0.53                           & 0.59                                                                         & 0.58                        \\
\bottomrule
\end{tabular}
\label{tab:mahalanobis-main}
\caption{Mean MD values for  all LLMs across all datasets. The range of Mean MD is [0,1] with zero being the best and one being the worst, except for Negation Dataset (for Negation Dataset one is the best and zero is the worst).}
\end{table*}

The embedding (representation) of each lexical item in an analogical pair (word embedding, sentence embedding) is extracted from eight LMs (In this work, we use the simplest representation, which is the \texttt{[CLS]} token representation). The distance measures between these two representations are then calculated using ED, CD, and MD.
For each dataset containing analogical pairs, these distance measures are calculated, and the mean of all the data points of a dataset is considered the representative distance for that dataset (these distances are Min-Max normalized).

Given the analogy taxonomy (figure \ref{analogy_tower}), except for the negation dataset at level 4, all the other datasets are positive analogies, meaning, that the two lexical items of a data point are considered analogical to each other. Therefore the mean distance values of these datasets should indicate such similarity (low cosine, Euclidean, and Mahalanobis distances). For the negation dataset, the two lexical items in a data point should not be analogical to each other. Therefore, the representative distance measures should be large. We discuss the implementation details in appendix \ref{sec:appendix-implementation}.

\section{Benchmark Results}

\subsection{Performance of LLMs on ANALOGICAL} 
We illustrate the performance of each LLM on different datasets at different levels of the analogy taxonomy based on the three distance measures in Figure \ref{fig:results-summary}. We further analyze the performance of LLMs based on MD akin to the superiority of MD over CD and ED mentioned in section \ref{sec:importance-md} (see Table 2). 
When inspecting the performance of LLMs at the word level, for BATS and MSR datasets, most LLMs perform considerably well with mean distance values close to zero. When moving into the word pair datasets (Google, SAT), all the LLMs struggle to perform with mean distance values closer to one. In word pair datasets, it is crucial to understand the implicit relations among the word pairs to model the analogies correctly in the vector space. The suboptimal performance exhibited by LLMs on the aforementioned datasets indicates the necessity of equipping them with the capability to identify implicit relationships. We believe that the integration of external knowledge into LLMs is a potential solution to enhance their performance on word pair analogies.

Analogies at level two (words vs. sentences) are also illustrated to be challenging for the LLMs to identify. These analogies are abstract since a single word represents the meaning of a sentence. Abstraction is an area of NLP that is yet to be studied systematically \cite{lachmy2022draw}. There are no widely established benchmarks to evaluate the performance of LLMs on abstraction. Therefore we postulate that it is hard for the LLMs to capture abstractions, performing poorly at this level.

The Random Reordering dataset is the hardest dataset for the LLMs at level three of analogy taxonomy compared to Random Deletion and Random Masking datasets. The current analogous sentences are created using a simple mechanism of deleting, reordering, or masking of words, as opposed to replacing nouns and/or verbs with their analogous counterparts. Therefore the resulting analogies should be easier for the LLMs to identify, as illustrated. 

At the fifth level, pertaining to entailment, the majority of LLMs demonstrate suboptimal performance, with the exception of T5, RoBERTa, and SpanBERT. Textual entailment consists of identifying semantically related sentences, and interpreting semantics is known to be a challenge to LLMs \cite{mayer2020enriching}, which explains the mean MD values closer to one.

Out of eight, six language models struggle to perform well at Metaphor Level. At this level, analogies are drawn between sentences and paragraphs, mainly introducing the issue of compositionality. Compositionality suggests that the meanings of complex expressions are constructed from the
meanings of the less complex constituents \cite{fodor2002compositionality}.
The inability of transformers to effectively capture the inherent compositionality in language, in the absence of suitable prompting techniques, has been extensively observed \cite{keysers2019measuring, furrer2020compositional}. We posit that this limitation directly contributes to the subpar performance of LLMs at this particular level.

\begin{figure*}[!ht]
\center
\includegraphics[width=.9\textwidth]{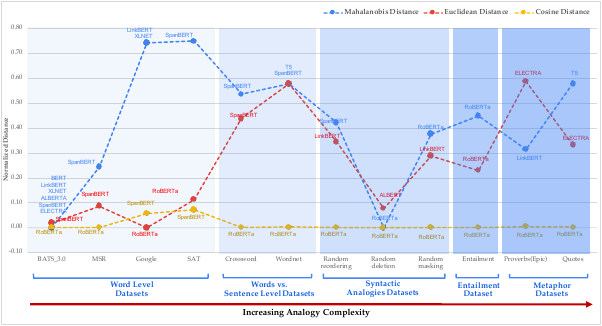}
\caption{Best performing model(s) for each dataset in each level of the analogy taxonomy (Performance on the Negation Dataset is shown separately in Figure \ref{fig:negation_dataset}). The range of each normalized distance measure is [0,1], with zero being the best and one being the worst.}
\label{fig:best_models}
\end{figure*}

\subsection{Performance on Negation Dataset}
\begin{figure}[!ht]
\center
\includegraphics[width=.9\columnwidth]{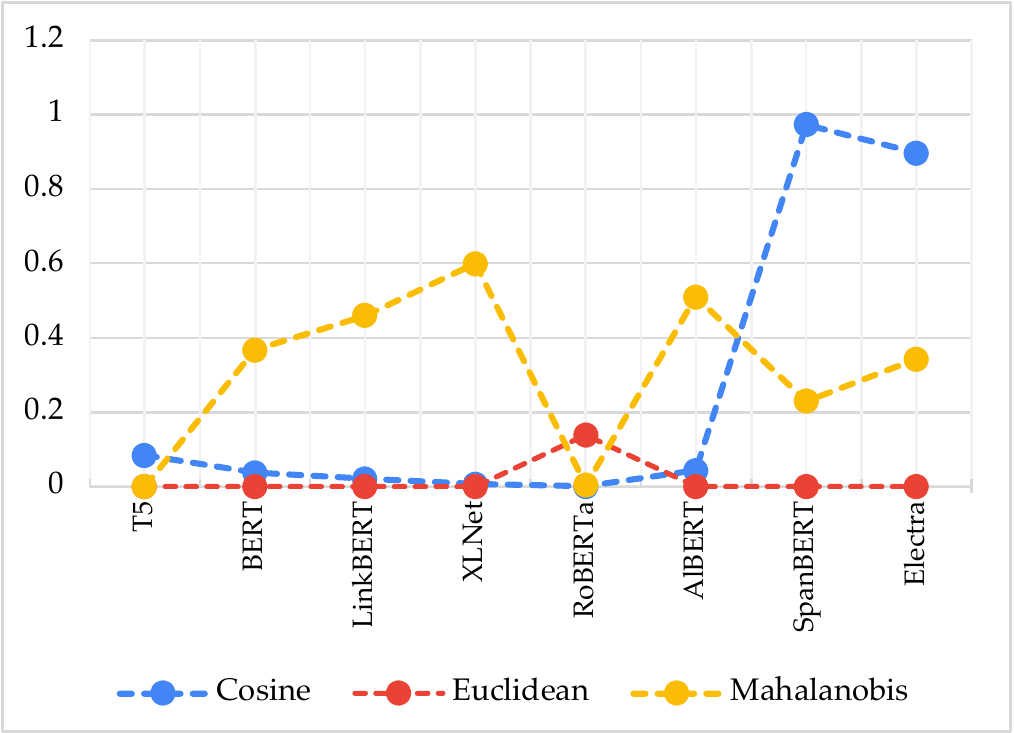}
\caption{Performance of LLMs on the Negation dataset. The range of each normalized distance measure is [0,1], with zero being the \textbf{worst} and one being the \textbf{best}.}
\label{fig:negation_dataset}
\end{figure}

Figure \ref{fig:negation_dataset} illustrates the performance of LLMs on the Negation Dataset. XLNET performs the best with a mean MD of 0.6. T5 and RoBERTa record the poorest performance by placing the negations pairs very closely in the vector space. This performance is justified based on previous research on negation identification by pretrained language models \cite{kassner-schutze-2020-negated}.

\subsection{Best performing LLMs}
In Figure \ref{fig:best_models}, we illustrate the best-performing models and their performance at each level of the analogy taxonomy across the three distance measures, ED, CD and MD. We see that RoBERTa performs the best based on mean CD values close to zero at all most all levels. However, CD considers all vector dimensions of a lexical item to be equally valuable and uncorrelated, which we reveal to be incorrect in section \ref{sec:importance-md}. Therefore we focus on the best-performing LLMs based on their mean MD values. We see that except for the Random Deletion dataset, the best performance for other datasets shows a general upward trend, indicating that it is increasingly hard for LLMs to identify analogous pairs when the complexity of the analogies increases. 
\vspace{-0.2em}

\section{Conclusion \& Future Avenues}
This work introduces ANALOGICAL, a benchmark for LLMs based on a taxonomy of six levels of analogies. Through comprehensive experiments, we show that LLMs increasingly struggle to identify analogies when the complexity of analogies increase (going up the analogy taxonomy). The datasets derived for level three are crude at this time. In the future, we will incorporate more challenging and comprehensive datasets to this level. We also will move on from this empirical study to investigate why some LLMs perform well at specific levels and not others.

\section{Limitations}

Syntactic analogies at level three consist of simple alterations of sentences based on deleting, reordering, and masking of random words. A more sophisticated method of creating syntactic analogies would be to replace nouns/ verbs in sentences with nouns and verbs of similar meaning, which is not explored in this work.


In this study, we utilize the \texttt{[CLS]} token as the representation for lexical items in analogies. While previous research efforts have investigated the optimal representations of lexical items in Large Language Models (LLMs) \cite{reimers-gurevych-2019-sentence, li-etal-2020-sentence}, we have chosen not to incorporate these findings into our current investigation.

This work uses mean distance measures to capture the LLMs' ability to identify analogies. There could be data points more challenging for the LLMs to capture than others within the same dataset or across datasets at the same level of the analogy taxonomy. Using mean distance values ignores this detail and considers all the data points equal, which is not optimum. 

\section*{Acknowledgements}
We thank Dr. Krishnaprasad Thirunarayan for his valuable feedback  and the anonymous reviewers for their constructive comments. This work was supported in part by the NSF grant \#2133842: EAGER: Advancing Neuro-symbolic AI with Deep Knowledge-infused Learning. Any opinions, findings, conclusions, or recommendations expressed in this material are those of the authors and do not necessarily reflect the views of the funding organization.

\bibliography{custom}
\bibliographystyle{acl_natbib}

\newpage
\appendix
\onecolumn

\section{Detailed Results}
\label{sec:appendix-detailed-results}
\begin{table}[!ht]
\centering
\scriptsize
\begin{tabular}{lrrrrrrrrrrrrr} 
\toprule
{\begin{tabular}[c]{@{}l@{}}Language \\Model\end{tabular}}  & \multicolumn{1}{l}{BATS\_3.0} & \multicolumn{1}{l}{MSR} & \multicolumn{1}{l}{Google} & \multicolumn{1}{l}{SAT} & \multicolumn{1}{l}{Crossword} & \multicolumn{1}{l}{Wordnet} & \multicolumn{1}{l}{\begin{tabular}[c]{@{}l@{}}Random \\Reordering\end{tabular}} & \multicolumn{1}{l}{\begin{tabular}[c]{@{}l@{}}Random \\Deletion\end{tabular}} & \multicolumn{1}{l}{\begin{tabular}[c]{@{}l@{}}Random \\Masking\end{tabular}} & \multicolumn{1}{l}{\begin{tabular}[c]{@{}l@{}}Stanford \\Negation\end{tabular}} & \multicolumn{1}{l}{Entailment} & \multicolumn{1}{l}{\begin{tabular}[c]{@{}l@{}}Proverbs\\(Epic)\end{tabular}} & \multicolumn{1}{l}{Quotes}  \\ 
\midrule
T5             & 0.04                          & 0.32                    & 0.95                       & 1.00                    & 0.60                          & 0.58                        & 0.51                                                                            & 0.22                                                                          & 0.41                                                                         & 0.00                                                                            & 0.48                           & 0.60                                                                         & 0.58                        \\
BERT           & 0.00                          & 0.38                    & 1.00                       & 0.97                    & 0.59                          & 0.60                        & 0.52                                                                            & 0.36                                                                          & 0.45                                                                         & 0.37                                                                            & 0.57                           & 0.63                                                                         & 0.62                        \\
LinkBERT       & 0.00                          & 0.55                    & 0.74                       & 1.00                    & 1.00                          & 0.96                        & 0.71                                                                            & 0.43                                                                          & 0.53                                                                         & 0.46                                                                            & 0.84                           & 0.31                                                                         & 0.98                        \\
XLNet          & 0.00                          & 0.55                    & 0.74                       & 1.00                    & 0.99                          & 0.99                        & 0.90                                                                            & 0.62                                                                          & 0.94                                                                         & 0.60                                                                            & 0.88                           & 0.32                                                                         & 1.00                        \\
RoBERTa        & 0.81                          & 0.82                    & 0.94                       & 0.95                    & 0.84                          & 0.90                        & 0.46                                                                            & 0.00                                                                          & 0.38                                                                         & 0.00                                                                            & 0.45                           & 1.00                                                                         & 0.92                        \\
AlBERT         & 0.00                          & 0.64                    & 0.96                       & 0.83                    & 0.97                          & 0.95                        & 0.83                                                                            & 0.46                                                                          & 0.74                                                                         & 0.51                                                                            & 0.71                           & 1.00                                                                         & 0.98                        \\
SpanBERT       & 0.00                          & 0.25                    & 1.00                       & 0.75                    & 0.54                          & 0.58                        & 0.42                                                                            & 0.23                                                                          & 0.39                                                                         & 0.23                                                                            & 0.49                           & 0.61                                                                         & 0.59                        \\
Electra        & 0.00                          & 0.26                    & 1.00                       & 0.82                    & 0.55                          & 0.58                        & 0.53                                                                            & 0.39                                                                          & 0.47                                                                         & 0.34                                                                            & 0.53                           & 0.59                                                                         & 0.58                        \\
\bottomrule
\label{tab:cosine}
\end{tabular}
\caption{Cosine Distance}
\end{table}

\begin{table}[!ht]
\centering
\scriptsize
\begin{tabular}{lrrrrrrrrrrrrr} 
\toprule
{\begin{tabular}[c]{@{}l@{}}Language \\Model\end{tabular}}  & \multicolumn{1}{l}{BATS\_3.0} & \multicolumn{1}{l}{MSR} & \multicolumn{1}{l}{Google} & \multicolumn{1}{l}{SAT} & \multicolumn{1}{l}{Crossword} & \multicolumn{1}{l}{Wordnet} & \multicolumn{1}{l}{\begin{tabular}[c]{@{}l@{}}Random \\Reordering\end{tabular}} & \multicolumn{1}{l}{\begin{tabular}[c]{@{}l@{}}Random \\Deletion\end{tabular}} & \multicolumn{1}{l}{\begin{tabular}[c]{@{}l@{}}Random \\Masking\end{tabular}} & \multicolumn{1}{l}{\begin{tabular}[c]{@{}l@{}}Stanford \\Negation\end{tabular}} & \multicolumn{1}{l}{Entailment} & \multicolumn{1}{l}{\begin{tabular}[c]{@{}l@{}}Proverbs\\(Epic)\end{tabular}} & \multicolumn{1}{l}{Quotes}  \\ 
\midrule
T5       & 0.56                          & 0.59                    & 0.71                       & 1.00                    & 0.85                          & 0.85                        & 0.77                                  & 0.31                                & 0.71                               & 0.00                                  & 0.55                           & 0.93                               & 0.77                        \\
BERT     & 1.00                          & 0.26                    & 0.32                       & 0.53                    & 0.49                          & 0.70                        & 0.51                                  & 0.22                                & 0.51                               & 0.00                                  & 0.57                           & 1.00                               & 0.68                        \\
LinkBERT & 0.63                          & 0.62                    & 0.98                       & 1.00                    & 0.75                          & 0.74                        & 0.35                                  & 0.18                                & 0.29                               & 0.00                                  & 0.47                           & 0.80                               & 0.82                        \\
XLNet    & 0.34                          & 0.30                    & 0.35                       & 0.74                    & 0.71                          & 0.85                        & 0.74                                  & 0.25                                & 1.00                               & 0.00                                  & 0.57                           & 0.74                               & 0.51                        \\
RoBERTa  & 0.37                          & 0.33                    & 0.00                       & 0.11                    & 0.52                          & 0.65                        & 0.53                                  & 0.26                                & 0.56                               & 0.14                                  & 0.46                           & 1.00                               & 0.70                        \\
AlBERT   & 0.50                          & 0.56                    & 0.14                       & 0.51                    & 0.88                          & 0.83                        & 0.48                                  & 0.08                                & 0.37                               & 0.00                                  & 0.23                           & 1.00                               & 0.91                        \\
SpanBERT & 0.02                          & 0.09                    & 0.24                       & 0.12                    & 0.44                          & 0.58                        & 0.40                                  & 0.19                                & 0.45                               & 0.00                                  & 0.46                           & 1.00                               & 0.78                        \\
Electra  & 0.43                          & 0.51                    & 0.50                       & 0.97                    & 1.00                          & 0.79                        & 0.76                                  & 0.32                                & 0.53                               & 0.00                                  & 0.37                           & 0.59                               & 0.33                        \\
\bottomrule
\label{tab:euclidean}
\end{tabular}
\caption{Euclidean Distance (Normalized)}
\end{table}

\section{Details on Distance measures}
\label{sec:appendix-detailed-distance}
\subsection{Euclidean Distance (ED)}
Euclidean distance is used to measure how far apart (in a straight line) two points are, in a vector space. If point $x$ and $y$ are represented in a higher dimensional vector space by $[x_1, \cdots, x_n]$ and $[y_1, \cdots, y_n]$ respectively, ED between $x$ and $y$ are given by:

\begin{align*}
    ED(x, y) &= \sqrt{\sum_{i=1}^{i=n} (x_i - y_i)^2}
\end{align*}

Values of ED range from $0$ to infinity. Zero indicates the two points are similar and larger numbers indicate the two points are far apart in the vector space and less similar.

\subsection{Cosine Distance (CD)}
Cosine similarity is a standard measure of similarity which measure the angle between two points in a vector space by taking into account the orientations of the vectors regardless of the vector sizes. Given points $U = [u_i, \cdots, u_n]$ and $V = [v_i, \cdots, v_n]$ in high-dimensional space cosine similarity between $u$ and $v$ is given by:

\begin{align*}
    CS(U, V) &= cos(\theta) &= \frac{\sum_{i=1}^{i=n} (u_iv_i)}{\sqrt{\sum_{i=1}^{i=n} u^2}\sqrt{\sum_{i=1}^{i=n} v^2}}
\end{align*}

We convert cosine similarity to cosine distance for easy comparison with Euclidian and Mahanalobis distances by subtracting cosine similarity from one.

\section{Details on Large Language Models}
\label{sec:appendix-detailed-llms}

\textbf{Bidirectional Encoder Representations from Transformers (BERT) \cite{devlin2018bert}} is trained on document-level corpora consisting of the BooksCorpus and English Wikipedia words through two unsupervised training tasks. In Masked Language Modeling (MLM) some tokens of input sequences are replaced randomly by a \texttt{[MASK]} token requiring BERT to predict the masked tokens allowing the LM to capture the directional nature of the language. To capture the relationships among sentences, BERT is trained on a second training objective known as Next Sentence Prediction (NSP).
 
\textbf{XLNet \cite{yang2019xlnet} }is a generalized autoregressive language model trained on corpora used by BERT as well as Giga5 (16GB text), ClueWeb 2012-B and Common Crawl corpora. XLNet improves upon BERT and introduces a permutation language modeling objective that retains benefits from both autoregressive and autoencoding pretraining objectives.
 
\textbf{RoBERTa \cite{liu2019roberta}} is as an optimized pretrained version of BERT, trained on a dataset ten times larger than BERT (16GB vs. 160GB) including the original dataset used to train BERT. In addition three other corpora containing news articles, web content, and a filtered subset of the CommonCrawl corpus were used. The training approach of RoBERTa differs from BERT as follows. RoBERTa modifies the MLM task by moving from static masking to dynamic masking where the masked tokens change at each epoch, thereby effectively leading to an increase in the diversity of learning opportunities for the model. RoBERTa removes NSP loss from the training objective arguing that the NSP loss was no longer required for better performance.
 
\textbf{A Lite BERT for Self-supervised Learning of Language Representations (ALBERT) \cite{lan2019albert}} targets to reduce the parameter size without affecting the performance of BERT. The LM is trained with the same corpora as BERT, yet three main changes to the BERT's design choices are made. The first change is feature factorization where input and hidden layers are decoupled from each other. Input vectors are first projected, to a lower dimensional embedding space and then into the hidden space, reducing the parameter size significantly. Secondly, parameters are shared across all layers (feed-forward and attention layers). Finally, ALBERT introduces a Sentence Order Prediction (SOP) loss in place of NSP, which is based on inter-sentence coherence. 

\textbf{Efficiently Learning an Encoder that Classifies Token Replacements Accurately (ELECTRA) \cite{clark2020electra}} introduces a new, more efficient training task aiming to reduce the computing power and retain or exceed the performance of previous BERT-based models pretrained on MLM task. ELECTRA's architecture includes two transformers, a generator, and a discriminator. The generator predicts the masked token from an input sequence and the resulting sequence is sent to the discriminator, which then predicts which tokens are original and which are predicted by the generator.

\textbf{SpanBERT \cite{joshi2020spanbert}} is specifically pretrained for improved predictions of spans of texts. SpanBERT introduces a new masking technique where spans of contiguous tokens are masked instead of individual tokens as in BERT. Also, the authors introduce a new training objective where the span boundary representations are used to predict
the entire content of the masked span.  

\textbf{Text-to-Text Transfer Transformer (T5) \cite{raffel2020exploring}} aims to introduce a unified framework for downstream NLP tasks. T5 is trained on the Colossal Clean Crawled Corpus (C4) introduced by the authors by removing text that is not natural language from the Common Crawl corpus. T5 has the vanilla encoder-decoder transformer architecture with an unsupervised training objective introduced by the authors inspired by MLM of BERT and word dropout regularization technique by \cite{bowman2015generating}.

\textbf{ A Knowledgeable Language Model Pretrained with Document Links (LinkBERT) \cite{yasunaga2022linkbert}} is an improvement over BERT, that incorporates document link knowledge into pretraining. The LM is trained on two joint objectives, MLM and Document Relation Prediction (DRP) and uses the same training dataset as BERT.

\section{Implementation Details}
\label{sec:appendix-implementation}

Hugging Face\footnote{\url{https://huggingface.co/models}} implementation of the LLMs (base configuration) are used to extract the word/sentence representations. In this study, we use the default configuration provided by Hugging Face (embedding size 768) if not specified otherwise. Scikit-learn\footnote{\url{https://scikit-learn.org/stable/index.html}} is used to implement the distance measures.

\end{document}